\documentclass[preprint,12pt]{elsarticle}

\usepackage{amssymb}

\usepackage{soul,color}
\usepackage{float}

\journal{Computer Methods and Programs in Biomedicine}

\begin{document}

\begin{frontmatter}



\title{Exploring deep learning methods for recognizing rare diseases and their clinical manifestations from texts}


\author[inst1]{Isabel Segura-Bedmar}
\author[inst1]{David Camino-Perdones}
\author[inst1,inst3,inst4,inst5]{Sara Guerrero-Aspizua}

\affiliation[inst1]{organization={Human Language and Accesibility Technologies, Computer Science Department, Universidad Carlos III de Madrid},
            addressline={Avenidad de la Universidad, 30}, 
            city={Leganés},
            postcode={28911}, 
            state={Madrid},
            country={Spain}}

\affiliation[inst2]{organization={Tissue Engineering and Regenerative Medicine group, Department of Bioengineering, Universidad Carlos III de Madrid},
            addressline={Avenidad de la Universidad, 30}, 
            city={Leganés},
            postcode={28911}, 
            state={Madrid},
            country={Spain}}

\affiliation[inst3]{organization={Hospital Fundación Jiménez Díaz e Instituto de Investigación, FJD},
            addressline={Av. de los Reyes Católicos, 2}, 
            city={Madrid},
            postcode={28040}, 
            state={Madrid},
            country={Spain}}

\affiliation[inst4]{organization={Epithelial Biomedicine Division, CIEMAT},
            city={Madrid},
            postcode={28040}, 
            state={Madrid},
            country={Spain}}
\affiliation[inst5]{organization={Centre for Biomedical Network Research on Rare Diseases (CIBERER)},
            addressline={C/Monforte de Lemos 3-5}, 
            city={Madrid},
            postcode={28029}, 
            state={Madrid},
            country={Spain}}

\begin{abstract}
Although rare diseases are characterized by low prevalence,
approximately 400 million people are affected by a rare disease. The early and accurate diagnosis of these conditions is a major challenge for general practitioners, who do not have enough knowledge to identify them. In addition to this, rare diseases usually show a wide variety of manifestations, which might make the diagnosis even more difficult. A delayed diagnosis can negatively affect the patient's life. Therefore, there is an urgent need to increase the scientific and medical knowledge about rare diseases. Natural Language Processing (NLP) and Deep Learning can help to extract relevant information about rare diseases to facilitate their diagnosis and treatments.
The paper explores several deep learning techniques such as Bidirectional Long Short Term Memory (BiLSTM) networks or deep contextualized word representations based on Bidirectional Encoder Representations from Transformers (BERT) to recognize rare diseases and their clinical manifestations (signs and symptoms). BioBERT, a domain-specific language representation based on BERT and trained on biomedical corpora, obtains the best results with an F1-score of 85.2\% for rare diseases.
\end{abstract}

\end{frontmatter}


\section{Introduction}
\label{sec:introduction}
Rare diseases are characterized by a low prevalence in the population. There is no consensus on the percentage of affected people with a disease to be considered as a rare disease. Thus, whereas in the United States, a rare disease affects fewer than 200,000 people, in Europe, the prevalence of a rare disease is less than 1 person per 2000   \cite{paz2010rare}. 
To date, there are around 7,000 rare diseases and new rare diseases are identified each week. In spite of their low prevalence, these diseases may affect more than 400 million people around the world  \cite{klimova2017global,ferreira2019burden}. 

The diagnostic process of rare diseases becomes a very long road for patients and their families to obtain an accurate diagnosis and then receive an adequate treatment. The delay in diagnosis of rare diseases is between six and seven years    \cite{zurynski2017australian}. A possible cause of the delayed diagnosis is the limited experience and knowledge about rare diseases of clinicians  \cite{ts2011general,domaradzki2019medical,elliott2015rare}. In addition, rare diseases may present a heterogeneous phenotype, with a wide variety of symptoms and signs, related among others with different driving mutations   \cite{moliner2010creating}. 
These characteristics are often responsible for inaccuracies in the diagnose of rare diseases. 
 Therefore, there is an urgent need to increase the usability of the sparse and fragmented  scientific and medical knowledge about rare diseases \cite{haendel2020many}. 

Artificial Intelligence, and in particular Natural Language Processing (NLP) and Machine Learning, can play a beneficial role by
providing better access to the relevant information about rare diseases and their clinical manifestations (signs and symptoms), and in this way, helping to alleviate the workload on doctors. Although much of the knowledge about rare diseases is stored in databases and ontologies, biomedical literature (research articles, clinical cases, health forums, social media, etc) is a rich source of information about rare diseases in unstructured text. Information extraction techniques such as Named Entity Recognition (NER) can help structure this information, facilitating  access to the knowledge embedded within those texts and boosting scientific research. 

The automatic recognition of disease named entities has attracted much attention over the last years  \cite{wei2016disease,habibi2017deep,xu2018sblc,zhao2017disease,ling2019domain,lee2020biobert,li2019fine}, as it can be applied in meaningful clinical applications such as cohort selection for clinical trials or epidemiological studies, pharmacovigilance, personalized medicine, among many others. 
This task is a very challenging task due to the diversity and complexity of disease names. Many disease names can have different synonyms and abbreviations to represent them. For instance, ``obsessive-compulsive disorder", ``obsessive compulsive disorder", ``anancastic neurosis", and ``OCD" are the same disease. Moreover, disease names usually contain modifiers that can be related to body parts or degrees of disease (e.g., ``periodic limb movement disorder" or ``advanced sleep phase syndrome"). 
The recognition of symptoms and signs also present additional challenges. 
Many symptoms  and signs can be described by technical terms (e.g., ``dysuria"), but also by short phrases (such as ``pain or discomfort when you urinate").
Furthermore, other NER challenges such as overlapping, nested and discontinuous entities have received limited attention  \cite{fei2021rethinking}.

The recent advancements of deep learning models have facilitated great progress in NLP. Recently, transformers  \cite{vaswani2017attention} and Bidirectional Encoder Representations from Transformers  \cite{devlin-etal-2019-bert} have outperformed traditional and  deep learning models for most of NLP applications \cite{yang2019xlnet,wu2019enriching,lewis-etal-2020-bart,zheng2019new}, and in particular, for NER in the biomedical domain \cite{lee2020biobert,hakala-pyysalo-2019-biomedical}. 

We briefly describe the most recent deep learning approaches for recognizing diseases in biomedical texts. One of the first studies that applied deep learning to this task is described in  \cite{wei2016disease}. The authors proposed a hybrid system composed of two modules: a  Conditional Random Field (CRF) \cite{lafferty2001conditional}  trained with orthographic, morphological, and domain features from Unified Medical Language System (UMLS) \cite{bodenreider2004unified},  and a bidirectional recurrent neural network (RNN) initialized with domain-specific word embeddings. Finally, a Support Vector Machine (SVM) classifier is used to combine the outputs of the two previous modules.  For the training and testing of the system, the authors used the dataset of the Disease Named Entity Recognition and Normalization (DNER) shared task~ \cite{wei2015overview} of the BioCreative V challenge, which consists of 1,500 PubMed abstracts and a total of 12,850 disease mentions. CRF achieves better results (F1=82.88\%) than the bidirectional RNN (F1=78.27\%). The output fusion by SVM obtains the best performance with an F1 of 84.28\%.

In the last years, Bidirectional Long Short Term Memory (BiLSTM) \cite{hochreiter1997long} with CRF has proved to be the most successful model for the task of biomedical NER~ \cite{lyu2017long,habibi2017deep,cho2019biomedical}. The approach proposed by Habibi et al.  \cite{habibi2017deep} was one of the first works to exploit pre-trained word embeddings to initialize a BiLSTM+CRF network for recognizing diseases. The authors used two pre-trained embedding models created by Pyysalo et al.  \cite{pyysalo2013dis}. The first model (from now on called PubMed-PMC) was trained using a collection of texts formed by all abstracts from PubMed (more than 23 million abstracts) and  all full articles from PMC (a database of open access with more than 700,000 full articles from the biomedical domain). The second embedding model (from now on called Wiki-PubMed-PMC) was an extension of the first one by adding approximately four million English articles from Wikipedia. These models were trained using the word2vec tool~ \cite{mikolov2013distributed}. The authors also trained a word embedding model by using a collection of 20,000 European patents. To train and evaluate their models, they use the NCBI corpus \cite{dougan2014ncbi} and the CDR corpus \cite{li2016biocreative}). 
The NCBI corpus is a collection of 793 PubMed abstracts and contains a total of 6,892 disease mentions. The CDR corpus contains 1500 MEDLINE abstracts annotated with diseases, chemicals, and their relations. The experiments showed that the network initialized with Wiki-PubMed-PMC obtains better performance (with an F1 of 90.4\% over the NCBI dataset and 88.17\% over the CDR dataset) than those initialized with the other pre-trained models. This may be because the Wiki-PubMed-PMC model was trained on a larger collection of texts than the other  pre-trained models. Moreover, this collections contained domain-specific and nonspecific texts. 

The SBLC model \cite{xu2018sblc}, is also based on a  BiLSTM network with a CRF layer. To represent the text, the authors trained a word embedding model by using a large collection of texts collected from PubMed, PMC,  and Wikipedia, with a total of  5.5 billion words. The SBLC was trained and tested on the NCBI dataset, obtaining an F1 of 86.2\%.

Instead of using RNN, Zhao et al.~ \cite{zhao2017disease} used a deep convolutional neural network (CNN). In addition to word embeddings, the authors also exploited character embeddings and lexicon feature embeddings to represent the texts. The character embeddings were generated by using a CNN layer. The MEDIC vocabulary  \cite{davis2012medic}, composed of more than 67,000 disease mentions, was used to create the lexicon feature embeddings. After the embedding layer, where each word is represented by concatenating its three embeddings, several CNN layers are applied to obtain higher level features. Then, instead of a CRF classifier, a multiple label strategy (MLS) is applied to capture the labels of the context words. This strategy uses a softmax function to obtain the probability of each possible label. The system obtained an F1 of 85.17\% on the NCBI corpus, and an F1 of 87.83\% on the CDR corpus.

Ling et al.  \cite{ling2019domain} also used an architecture composed of a BiLSTM with a CRF layer. This architecture was  initialized by using the three type of embeddings proposed by Zhao et al. ~ \cite{zhao2017disease}, as just described above. The main difference is that these authors applied a combination of a CNN and a LSTM to generate the character embeddings, instead of using a CNN network. The final model achieved an F1 of 83.8\% on the NCBI dataset. 


One of the main drawbacks of the pre-trained word embeddings models is that they only provide a vector for each word, so they do not handle polysemous words. 
Recently, contextualized word representation models (such as ELMo \cite{peters-etal-2018-deep}, GPT-2 \cite{radford2019language} or BERT \cite{devlin-etal-2019-bert}) have emerged as an alternative to the non-contextual word embedding models, providing a different vector for each sense of a word.
Lee and colleagues  \cite{lee2020biobert} applied  BERT to the task of disease recognition on the NCBI dataset, achieving an F1 of 88.60\%. The authors also trained their language representation model (BioBERT) on two large biomedical corpora such as PubMed and PMC. BioBERT slightly overcomes BERT on the NCBI dataset, with an improvement of 0.62\%. 

Li et al.  \cite{li2019fine} also trained a BERT model using 1.5 million electronic health record notes. This model was evaluated on the NCBI and CDR datasets, showing an F1 of 89.92\% and 93.82\% respectively. 

Very few research efforts have focused on the extraction of rare diseases. The RDD corpus \cite{fabregat2018deep} contains 1000 MedLine abstracts covering 578 rare diseases and 3,678 annotations expressing a
disability. A disability  can be defined as ``any restriction or lack (resulting
from an impairment) of ability to perform an activity in the manner or within the
range considered normal for a human being"\cite{world1980international}. The authors analyzed a model based on Bi-LSTM and CRF to extract 
rare diseases and disabilities, achieving an F1-score of 70.1\% for rare diseases and 81\% for disabilities. 

In this paper, we address the task of recognizing rare diseases as well as their clinical manifestations (symptoms and signs). Moreover, to the best of our knowledge, this is the first work 
 that explores three BERT-based models to extract rare diseases from texts. 
In particular, we use the basic BERT model and two models, BioBERT \cite{lee2020biobert}, and ClinicalBERT \cite{alsentzer-etal-2019-publicly}, which were trained using biomedical and clinical texts, respectively.  In order to provide a comprehensive comparison, we also study  several BiLSTM models initialized with different pre-trained word embedding models.

\section{Methods}
\label{sec:approach}

\subsection{Dataset}
We use the RareDis corpus \cite{martinez2021RareDis}, which is a collection of texts from the Rare Disease database (NORD)\footnote{https://rarediseases.org/}. These texts were manually annotated with four entity types (diseases, rare diseases, signs, and symptoms). The corpus also includes relations between entities, but they are outside the scope of this work. 
The corpus has three different splits: training set, validation set, and test set.  
Table \ref{tab:raredis} shows the number of the entity types annotated, as well as the number of documents, sentences, and tokens in each split. 
A more detailed description of the RareDis corpus can be found in  \cite{martinez2021RareDis}. The corpus contains a total of 9,318 entities. We can observe that sign and rare disease entity types are the most prevalent, around 41\% and 34\%, respectively. The disease entity type is the third-largest type, with approximately 17\%, while symptom entity type is the most sparse entity type in the three splits.

\begin{table}[!htbp]
\centering
\caption{Statistics of the RareDis corpus.}
      \begin{tabular}{lcccc}
        \hline
           & Training  &Validation   & Test & Total\\ \hline
           
        Documents & 729 & 104 & 208 & 1,041\\

        Sentences & 6,451 & 903  & 1,787 & 9,141\\
        Tokens & 135,656  & 18,492   & 37,893 & 192,041\\\hline 
        Diseases & 1,647 & 230 & 454 & 2,331\\
        Rare Diseases & 3,608 & 525  & 1,095 & 5,228\\
        Symptoms & 319  & 24   & 54 & 397\\ 
        Signs & 3,744  & 528   & 958 & 5,230\\ 
        \hline 
        
      \end{tabular}
      \label{tab:raredis}
\end{table}

The corpus is distributed in Brat standoff format  \cite{stenetorp2012brat}. We parse the texts using Spacy to obtain their tokens, lemmas, and PoS tags. As NER is a  sequence labeling task, we represent each token using the standard IOB2 (Inside, Outside, Beginning) encoding scheme  \cite{tjong-kim-sang-veenstra-1999-representing}, where  
 B-X identifies the first token of an entity mention whose type is X (for example, B-SIGN), I-X identifies the continuation of an entity mention with type X (for example, I-SIGN), and O for other tokens. 
 The RareDis corpus and its guidelines are publicly available for the research community\footnote{https://github.com/isegura/NLP4RARE-CM-UC3M}.

\subsection{Approaches}

Now we describe the different methods used to deal with the task of NER on the RareDise corpus. 

As a baseline system for comparison, we use a CRF, one of the most successful algorithms for any sequence labeling task such as NER \cite{nguyen2007comparisons,li2020survey}.
Moreover, we explore the use of different deep learning architectures such as BiLSTM and Transformers. 

For much of the past decade, RNNs, and in particular BiLSTM, have been successfully applied to a wide range of NLP tasks \cite{peters-etal-2018-deep,DBLP:conf/coling/ZhouQZXBX16}. However, these networks are characterized by some drawbacks such as difficulty to process longer sequences and high computational complexity \cite{vzukov2018named}. On the contrary, transformers are based on attention mechanism \cite{luong-etal-2015-effective}, which are capable to capture the relevant information of the input sequence and also allows parallelization (it can be executed in parallel for each word in the input sequence, while RNNs have to process word by word). 
Moreover, transformers have outperformed RNNs, setting the new state-of-the-art performance in many NLP tasks \cite{wolf-etal-2020-transformers}.

Thus, we aim to provide a comparative analysis of deep learning models for detecting rare diseases and their clinical manifestations in texts.

\subsubsection{CRF}

CRF learns the correlations between labels and provides the output sequence of IOB tags with the highest probability. As the feature set, we consider three kinds of features: token, lemma, and PoS tag. For each token, we select a window of size two. Then, the features of the tokens belonging to this window are the representation of each token. These features are fed into the CRF classifier, which predicts an IOB tag for each input token. To implement the model, we use the CRFSuite package \cite{okazakiCRFsuite2007}. The classifier was trained using both training and validation datasets since we use default hyperparameters. The Limited Memory Algorithm for Bound Constrained Optimization (L-BFGS) is used as the optimization method.

\subsubsection{Bidirectional Long short-term memory (BiLSTM)}

BiLSTM has been successfully applied to the NER task in the biomedical domain \cite{lyu2017long,zeng2017lstm}, and in particular, to recognize disease names \cite{wei2016disease,habibi2017deep,xu2018sblc,ling2019domain}. This model consists of a forward LSTM (which sequentially processes the input sequence from left to right) and a backward LSTM (which processes the input sequences from right to left). In this way, BiLSTM can learn relevant information from the previous and next context for each input token, effectively increasing the amount of information available to the network  \cite{1556215}.

Our architecture consists of several layers, which are described below. First, in the input layer, the text is represented as word vectors. Then, these input vectors are passed to the BiLSTM layer described above. The output vector of the BiLSTM layer is the concatenation of the forward LSTM and the backward LSTM.
After the BiLSTM layer, we consider two different strategies for the output layer. 
The first strategy is using a CRF classifier as the last layer, which will output the sequence of IOB tags with the maximum probability for the input sequence. The CRF layer takes as input the label probability for each word coming from the output layer of the BiLSTM model. Thus, the context surrounding the label assignment predicted by the BiLSTM model is also added, whereby linear-chain CRF explicitly models dependencies between the labels through a transition matrix with transition scores between all pairs of the labels. This allows to easily learn constraints such as, for example,  "I-RAREDISEASE" tag cannot follow an "O" tag. These types of constraints are captured by the CRF layer in a simple way by considering the time step in each token. As a second strategy, we also evaluate a BiLSTM without CRF layer, where each probability is treated conditionally independent. To do this, instead of using CRF, we employ a TimeDistributedDense layer, very similar to a deep layer that can be applied to every time-step of the BiLSTM layer.

Moreover, we explore the effect of input text representation on the performance of BiLSTM. Texts must be encoded as vectors of real numbers to be used as input for machine learning and deep learning models. In the case of neural networks, it is possible to create a random vector for each input token. During the training, the network will adjust these word vectors alongside the other weights of the network.
An alternative way is to represent tokens with word vectors (word embeddings) from a language model. 
In the last decade, neural network language models \cite{mikolov2012context,arisoy2012deep} have effectively replaced traditional models such as the Bag-Of-Words, achieving state-of-the-art results in many NLP tasks. Several studies have shown that word embeddings trained with neural networks can capture semantic and syntactic between tokens \cite{mikolov2013distributed}, providing thus an accurate meaning representation of the input tokens. The most popular word embeddings models are Word2Vec \cite{mikolov2013distributed}, Glove \cite{pennington-etal-2014-glove}\ and fastText \cite{bojanowski2017enriching}. In this work, we study the effect of different pre-trained word embeddings on the BiLSTM performance. In particular, we explore three different models: 
\begin{itemize}
    \item GoogleNews \cite{NIPS2013_9aa42b31}, a pre-trained word embedding model trained with the Word2Vec network on the GoogleNews dataset. The model contains word embeddings of dimension 300 for 3 million words.
    \item GloVe \cite{pennington-etal-2014-glove}, a pre-trained word embedding model trained using Common Crawl, an open repository of web crawl data. The model contains 300-dimensional vectors for 840 billion tokens. 
    \item PubMed, PubMed Central, and Wikipedia (Wiki-Pubmed-PMC) \cite{pyysalo2013we}, a pre-trained word embedding model trained with the Word2Vec network on a collection of more than 23 million abstracts from PubMed (a database containing abstracts of scientific articles from the biomedical domain), 700,000 articles from PMC and around four million English Wikipedia articles. The dimension of the word embeddings is 200. 
    
\end{itemize}

To implement and train the models, we use the Keras Python API \cite{keras} with TensorFlow as the backend.
 We use an Adam optimizer \cite{kingma2017adam} with a learning rate 0.001 and categorical cross-entropy as a loss function. To avoid overfitting, we use early stopping with the patience of four, meaning that training will finish if the loss function does not improve in four consecutive epochs.

\subsubsection{Bidirectional Encoder Representations from Transformers (BERT)}

Deep contextualized language models are capable to capture  word meanings and their more representative relations with other words. Thanks to this accurate linguistic representation, these models achieved unprecedented results on many NLP tasks \cite{devlin-etal-2019-bert}. 
 Moreover, contextualized language models are trained through unsupervised learning, requiring only a plain text corpus. Thus, these models can partially alleviate the shortage of large annotated corpora, which are essential for supervised machine learning algorithms. 

Without a doubt, BERT, which stands for Bidirectional Encoder Representations from Transformers, is the most popular contextualized language model due to its excellent results in many NLP applications \cite{devlin-etal-2019-bert}. Transformers are based on attention mechanism \cite{vaswani2017attention}, which attempts to represent each word in a sentence based on the most relevant tokens for that word. Attention mechanisms present two major advantages compared with RNN: first, these mechanisms can handle long-term dependencies between any two tokens in a sentence, and second, they can enable the parallelization of training.

The basic idea of BERT is that the model is trained to predict words from their contexts in an unsupervised way. This prediction only requires a large collection of texts and some strategy to mask those words to be predicted.  Thus, BERT can learn meaningful representation for the words in a sequence. The architecture of BERT (which consists of 12 encoder layers for the BERT-base version or 24 encoder layers for the BERT-large variation) can be extended with more layers capable to solve a specific NLP task. 
This process is known as fine-tuning. 
In our case, we have used the BertForTokenClassification class provided by the PyTorch-Transformers package\footnote{https://pytorch.org/hub/huggingface\_pytorch-transformers/}, which is a library of state-of-the-art pre-trained models for NLP. It provides PyTorch implementations and pre-trained model weights for the most popular deep contextualized language models.
The BertForTokenClassification class implements a fine-tuning model that adds a token-level classifier on top of the BERT model. The token-level classifier is a linear layer that takes as input the last hidden state of the sequence. 
The BertForTokenClassification class  allows to load different pre-trained model. In this work, we explore the following  ones: 
\begin{itemize}
    \item Bert-base-uncased version of the original BERT proposed in  \cite{devlin-etal-2019-bert}. This version is a stack of 12 encoders, each having 12 attention heads.  For each token of the input sentence, the output layer provides an embedding of dimension 768 for this token. The total number of parameters is 110 million. The model was trained using two corpora: BookCorpus with around 800 million words and English Wikipedia with around 2,500 million words.
    
    \item BioBERT  \cite{lee2020biobert}, whose weights were initialized using the BERT weights, and then, the model was pre-trained on two biomedical corpora: PubMed abstracts (4,500 million words) and PMC full-text articles (13,500 million words). 
    \item ClinicalBERT  \cite{alsentzer-etal-2019-publicly} was trained with more than 2 million clinical notes from the MIMIC-III v1.4 database \cite{johnson2016mimic}. Its weights were initialized using the BioBERT weights. 
    
\end{itemize}

\section{Results and discussion}
\label{sec:eval}

In this section, the  results obtained from the different methods are presented. 
To calculate the evaluation metrics (accuracy, recall, precision and F1-score), the sklearn-crfsuite\footnote{https://sklearn-crfsuite.readthedocs.io/en/latest/} and seqeval \cite{seqeval} packages libraries  are used to calculate the results at the token and entity level, respectively.

Firstly, we start by presenting the micro-average F1 of the approaches used in this study (Table \ref{tab:Summaries}). We can clearly see that the BERT-based models outperforms all the other models. 
Clearly, the deep contextualised vectors from the BERT-based models provide a better representation for the input texts than those provided by CRF or the pre-trained word embeddings used in BiLSTM. BioBERT obtains better results than BERT and ClinicalBERT. This may happen because this was trained on biomedical scientific articles, whose narrative is similar to that used in the NORD database for describing  rare diseases. Regarding the other approaches, although BiLSTM was extended with a CRF layer as the output layer, this architectures does not obtain better results than a simple CRF. A possible reason could be  
that this deep learning technique requires a larger number of training examples for learning. 

\begin{table}[!htbp]
\centering
\small
\caption{\label{tab:Summaries}Comparison of the methods}
\begin{tabular}{lc}
\\
{\bf Approach}&
{\bf F1-score} \\
\hline\\[-8pt]

CRF &	0.6487	\\
BiLSTM (Wiki-PubMed-PMC) &	0.4326\\
BiLSTM+CRF (Wiki-PubMed-PMC)		&	0.5805	\\
BERT  & 0.6710 \\
BioBERT  & \bf 0.6954 \\
ClinicalBERT  & 0.6810\\

\hline
\end{tabular}
\end{table}

We discuss the results of each of these approaches in more detail below.

\subsection{CRF (baseline)}

Table \ref{tab:CRFentity} shows results achieved by CRF on entity-level. Token-level results are shown in \ref{sec:apptoken}. CRF achieves a micro-average F1-score of 64.8\% and a macro-average F1-score of 61.9\%. As classes are unbalanced, we also consider the macro-weighted-average F1-score, which is of 63.8\%.

\begin{figure}
  \caption{Examples of entity types in the RareDis corpus.}
      \includegraphics[scale=0.40]{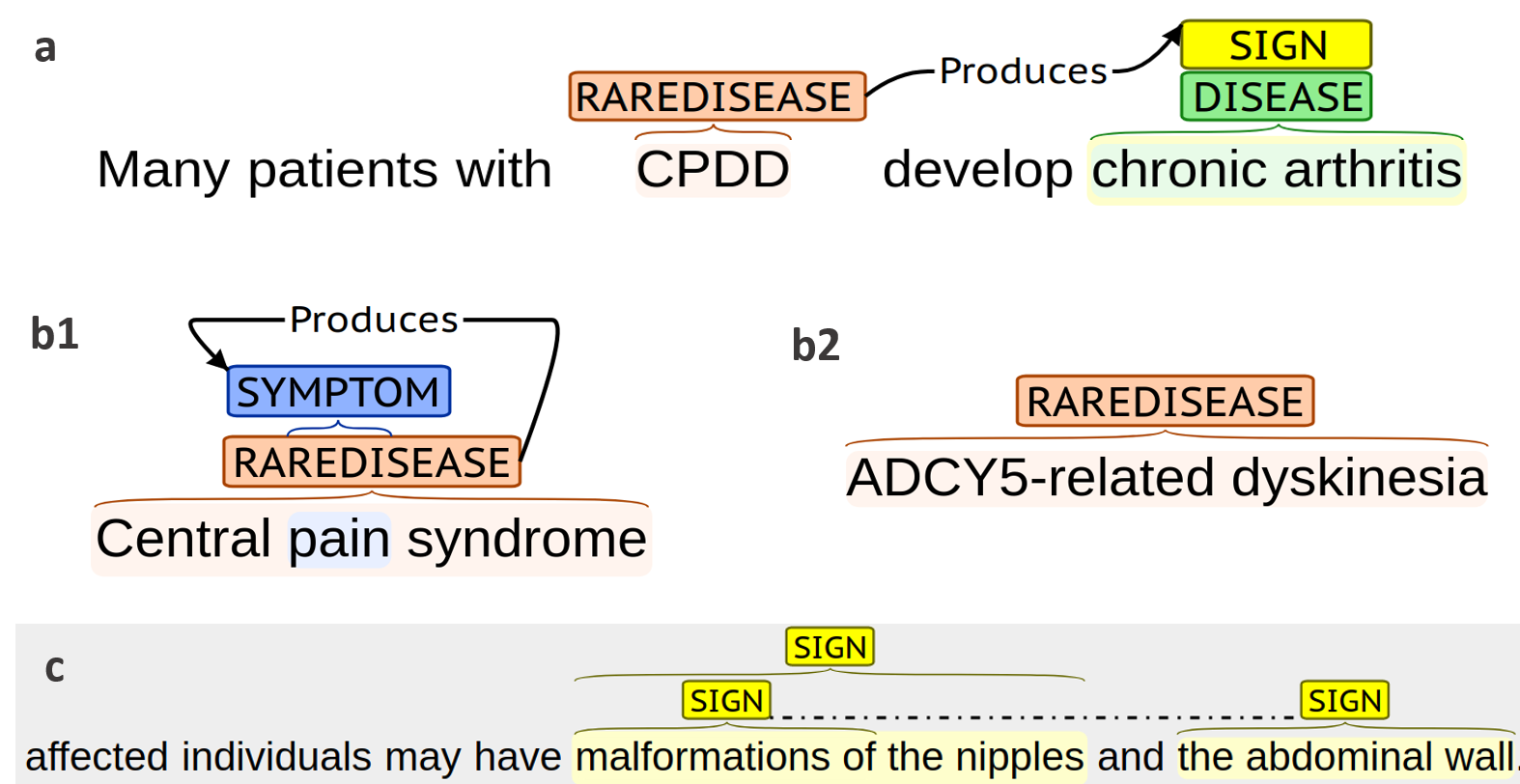}
     \label{fig:examples}
    \end{figure}

The best results are obtained for rare disease entity type (F1=82.4\%), followed by symptom  (F1=62.2\%). 
On the contrary, sign entity type shows the lowest F1-score (45.5\%) value, despite being the entity type with the largest number of instances (41\%) in the training dataset (see Table \ref{tab:raredis}). 
This may happen because sign mentions are usually nominal phrases (for example, ``malformations of the nipples"), unlike disease or rare disease names, which are usually a combination of few technical terms (for example,  ``ADCY5-related dyskinesia"). In Table \ref{tab:CRFtoken}, the ``Support" column shows the number of instances for each type of token. The number of internal tokens (I-) for diseases or rare diseases is slightly higher than the number of its initial tokens (B-), while the number of internal tokens for signs doubles the number of its initial tokens. In addition, many sign mentions are discontinuous entities, that is, they present gaps in their description. The sentence shown in Figure \ref{fig:examples}.c contains two signs: ``malformations of the nipples" and ``malformations of the abdominal wall", being the last one a discontinuous mention.
Another possible reason is that many signs can be also considered as diseases (see Figure \ref{fig:examples}.a). CRF and the other models proposed in this study only provide a label per token. That is, they do not address the task of overlapped entities.  

\begin{table}[!htbp]
\centering
\small
\caption{\label{tab:CRFentity}Entity-level results of CRF. Best scores are in bold.}
\begin{tabular}{lcccc}
\\
{\bf Label} & {\bf Precision} &
{\bf Recall} &
{\bf F1-score} & {\bf Support} \\
\hline\\[-8pt]

DISEASE	&	0.6991	&	0.4912	&	0.5770	&	454\\
RAREDISEASE	&	0.8332	&	0.8164	&	0.8247	&	1095\\
SIGN	&	0.5313	&	0.3987	&	0.4556	&	958\\
SYMPTOM	&	0.7778	&	0.5185	&	0.6222	&	54\\
micro-avg	&	0.7112	&	0.5963	&	0.6487	&	2561\\
macro-avg	&	0.7103	&	0.5562	&	0.6199	&	2561\\
macro-weighted	&	0.6953	&	0.5963	&	0.6384	&	2561\\
\hline
\end{tabular}
\end{table}

Both signs and symptoms are clinical manifestations of diseases. A sign is an objective evidence, while a symptom is a subjective experience that can only be identified by the patient. However, contrary to the low results for signs, CRF provides the second-best F1-score for symptom type, which has the lowest number of instances (see Table \ref{tab:raredis}). A manual review of symptoms and signs mentions in the training dataset shows that most symptoms are described by technical terms (for example, ``headache"), while signs usually have lay descriptions (for example, ``dark circles under eyes"). It would be necessary to increase the number of symptoms in the RareDis corpus to study whether the difference between the results of both types of entities is maintained.

 \subsection{BiLSTM}

All the BiLSTM models provide significantly lower  results than CRF (see Tables  \ref{tab:CRFentity} and \ref{tab:BiLSTM}).
The decrease in micro-average F1-score is more than 20\% and 24\% in macro-average F1-score. 
This may indicate that the training data is too small for using deep learning. 
As happened with CRF, BiLSTM obtains the best results for rare diseases and worst ones for signs. 

Regarding the effect of pre-trained word embeddings to initialize the network, the BiLSTM with 
Wiki-Pubmed-PMC provides the best overall results. It also obtains the best results for rare diseases and diseases. This may be because these word embeddings were trained on biomedical texts. 
BiLSTM with Glove achieves a slightly better F1-score for signs than BiLTM with Wiki-Pubmed-PMC. However, BiLSTM with Glove achieves an improvement of almost 6\% of F1-score for symptoms over BiLSTM with Wiki-Pubmed-PMC. Although Glove word embeddings were not trained on biomedical texts, they obtain very close results to those obtained with Wiki-Pubmed-PMC. This may be because Glove has the biggest vocabulary size. 
On the other hand, random initialization and GoogleNews word embeddings provide lower results.

\begin{table}[!htbp]
\centering
\small
\caption{\label{tab:BiLSTM}Entity-level results of  BiLSTM models. Best scores are in bold.}
\begin{tabular}{lcccc}
\\
& \multicolumn{1}{c}{\textbf{Random Initialization}} \\
{\bf Label} & {\bf Precision} &
{\bf Recall} &
{\bf F1-score} & {\bf Support} 
\\
\hline\\[-8pt]

DISEASE	&	0.4387	&	0.2913	&	0.3502	&	454\\
RAREDISEASE	&	0.4592	&	0.4712	&	0.4651	&	1095\\SIGN	&	0.3288	&	0.3224	&	0.3256	&	958\\SYMPTOM	&	0.0000	&	0.0000	&	0.0000	&	54\\

micro-avg	&	0.3668	&	0.3742	&	0.3705	&	2561\\macro-avg	&	0.2454	&	0.2170	&	0.2282	&	2561\\
macro-weighted	&	0.3946	&	0.3742	&	0.3820	&	2561\\
\hline

& \multicolumn{1}{c}{\textbf{Google News}} \\
{\bf Label} & {\bf Precision} &
{\bf Recall} &
{\bf F1-score} & {\bf Support} \\
\hline\\[-8pt]

DISEASE	&	0.4432	&	0.3071	&	0.3628	&	454\\RAREDISEASE	&	0.4796	&	0.4971	&	0.4882	&	1095\\SIGN	&	0.3166	&	0.3419	&	0.3287	&	958\\SYMPTOM	&	0.4571	&	0.3200	&	0.3765	&	54\\
micro-avg	&	0.3724	&	0.4020	&	0.3866	&	2561\\macro-avg	&	0.3393	&	0.2932	&	0.3112	&	2561\\
macro-weighted	&	0.4084	&	0.4020	&	0.4028	&	2561\\

\hline
& \multicolumn{1}{c}{\textbf{Glove}} \\
{\bf Label} & {\bf Precision} &
{\bf Recall} &
{\bf F1-score} & {\bf Support} \\
\hline\\[-8pt]
DISEASE	&	0.4246	&	0.3622	&	 0.3909	&	454\\RAREDISEASE	&	0.5194	&	0.5529	&	0.5356	&	1095\\SIGN	&	0.3114	&	0.3971	&	\bf 0.3491	&	958\\
SYMPTOM	&	0.6154	&	0.4800	&	\bf 0.5393	&	54\\micro-avg	&	0.3850	&	0.4596	&	0.4190	&	2561\\macro-avg	&	0.3742	&	0.3584	&	0.3630	&	2561\\
macro-weighted	&	0.4236	&	0.4596	&	0.4387	&	2561\\
\hline
& \multicolumn{1}{c}{\textbf{Wiki-Pubmed-PMC}} \\
{\bf Label} & {\bf Precision} &
{\bf Recall} &
{\bf F1-score} & {\bf Support} \\
\hline\\[-8pt]
DISEASE	&	0.5794	&	0.4339	&	\bf 0.4962	&	454\\
RAREDISEASE	&	0.5378	&	   0.5388	&	\bf 0.5383	&	1095\\
SIGN	&	 0.3167	&	0.3570	&	 0.3356	&	958\\
SYMPTOM	&	0.5946	&	0.4074	&	 0.4835	&	54\\
micro-avg	&	0.4170	&	0.4494	&	\bf 0.4326	&	2561\\macro-avg	&	0.4057	&	0.3474	&	\bf 0.3707	&	2561\\
macro-weighted	&	0.4637	&	0.4494	& \bf	0.4539	&	2561\\

\hline
\end{tabular}
\end{table}

\subsection{BiLSTM-CRF}

Table \ref{tab:bilstmcrf} shows the results obtained by the BiLSTM-CRF. In all the BiLSTM-CRF models, the CRF layer helps  outperform the same models without using CRF, with improvements around 10-15\% over the BiLSTM overall scores. 
However, BiLSTM-CRF models still provide lower overall results than the baseline based on CRF, with a decrease of 6\% in micro-average F1-score. 

The BiLSTM-CRF with Wiki-Pubmed-PMC word embeddings achieves the best overall results. Moreover, this model also provides the best F1-scores for diseases and  symptoms, while the BiLSTM-CRF with Glove provides the best results for rare diseases and signs. 
The BiLSTM-CRF initialized with random vectors or GoogleNews word embeddings have  show similar results. They do not outperform the BiLSTM+CRF with  Wiki-Pubmed-PMC or Glove.

\begin{table}[!htbp]
\centering
\small
\caption{\label{tab:bilstmcrf}Entity-level results of BiLSTM-CRF models. Best scores are in bold.}
\begin{tabular}{lcccc}
\\
& \multicolumn{1}{c}{\textbf{Random Initialization}} \\
{\bf Label} & {\bf Precision} &
{\bf Recall} &
{\bf F1-score} & {\bf Support} \\
\hline\\[-8pt]

DISEASE	&	0.5414	&	0.3780	&	0.4451	&	454\\
RAREDISEASE	&	0.6540	&	0.7144	&	0.6829	&	1095\\
SIGN	&	0.4892	&	0.4391	&	0.4628	&	958\\
SYMPTOM	&	0.8529	&	0.5800	&	0.6905	&	54\\
micro-avg	&	0.5421	&	0.5494	&	0.5457	&	2561\\
macro-avg	&	0.5075	&	0.4223	&	0.4563	&	2561\\
macro-weighted	&	0.5748	&	0.5494	&	0.5582	&	2561\\
\hline

& \multicolumn{1}{c}{\textbf{Google News}} \\
{\bf Label} & {\bf Precision} &
{\bf Recall} &
{\bf F1-score} & {\bf Support} \\
\hline\\[-8pt]

DISEASE	&	0.5597	&	0.4304	&	0.4866	&	454\\
RAREDISEASE	&	0.6482	&	0.7548	&	0.6975	&	1095\\
SIGN	&	0.5327	&	0.4166	&	0.4675	&	958\\
SYMPTOM	&	0.6667	&	0.5600	&	0.6087	&	54\\
micro-avg	&	0.5556	&	0.5654	&	0.5604	&	2561\\
macro-avg	&	0.4815	&	0.4324	&	0.4521	&	2561\\
macro-weighted	&	0.5887	&	0.5654	&	0.5711	&	2561\\

\hline

& \multicolumn{1}{c}{\textbf{Glove}} \\
{\bf Label} & {\bf Precision} &
{\bf Recall} &
{\bf F1-score} & {\bf Support} \\
\hline\\[-8pt]
DISEASE	&	0.4720	&	0.5092	&	0.4899	&	454\\
RAREDISEASE	&	0.7226	&	0.7240	&	\bf 0.7233	&	1095\\
SIGN	&	0.5068	&	0.4606	&	\bf 0.4826	&	958\\
SYMPTOM	&	0.5385	&	0.5600	&	0.5490	&	54\\
micro-avg	&	0.5489	&	0.5821	&	0.5650	&	2561\\
macro-avg	&	0.4480	&	0.4508	&	0.4490	&	2561\\
macro-weighted	&	0.5937	&	0.5821	&	0.5874	&	2561\\

\hline

& \multicolumn{1}{c}{\textbf{Wiki-Pubmed-PMC}} \\
{\bf Label} & {\bf Precision} &
{\bf Recall} &
{\bf F1-score} & {\bf Support} \\
\hline\\[-8pt]
DISEASE	&	0.7208	&	0.4890	&	\bf 0.5827	&	454\\
RAREDISEASE	&	0.6339	&	0.7890	&	0.7030	&	1095\\
SIGN	&	0.4994	&	0.4562	&	0.4768	&	958\\
SYMPTOM	&	0.6739	&	0.5741	&	\bf 0.6200	&	54\\
micro-avg	&	0.5564	&	0.6068	&	\bf 0.5805	&	2561\\
macro-avg	&	0.5056	&	0.4617	&	\bf 0.4765	&	2561\\
macro-weighted	&	0.5998	&	0.6068	&	\bf 0.5953	&	2561\\

\hline
\end{tabular}
\end{table}

As mentioned previously, BiLSTM fails to beat the baseline, not even when it includes a CRF classifier as its last layer. This may be because the training data size is not enough to train a deep learning model, while a CRF classifier trained with a simple feature set can deal with the task. 
Regarding the pre-trained word embeddings, Wiki-Pubmed-PMC and Glove word embeddings provide better performance than using random initialization or GoogleNews word embeddings. 

\subsection{BERT-based models}
We have explored the use of three different deep contextualized word representations, all of them based on BERT (see Table \ref{tab:bert}). Unlike the BiLSTM models, these BERT-based models exceed the baseline results provided by a simple CRF classifier. 

\begin{table}[!htbp]
\centering
\small
\caption{\label{tab:bert}Entity-level results of the BERT-based models. Best scores are in bold.}
\begin{tabular}{lcccc}
\\
& \multicolumn{1}{c}{\textbf{BERT base}} \\
{\bf Label} & {\bf Precision} &
{\bf Recall} &
{\bf F1-score} & {\bf Support} \\
\hline\\[-8pt]
DISEASE & 0.5197    &0.6101 & 0.5613 & 454\\
RAREDISEASE & 0.8008 & 0.8667 & 0.8325 & 1095\\
SIGN & 0.5079 & 0.6033 & 0.5515 & 958\\
SYMPTOM & 0.5469 & 0.6481 & 0.5932 & 54\\

micro avg & 0.6298 & 0.7181 & 0.6710 & 2561\\
macro avg & 0.5938 & 0.6821 & 0.6346 & 2561\\
macro-weighted & 0.6361 & 0.7181 & 0.6743 & 2561\\

\hline

& \multicolumn{1}{c}{\textbf{BioBERT}} \\
{\bf Label} & {\bf Precision} &
{\bf Recall} &
{\bf F1-score} & {\bf Support} \\
\hline\\[-8pt]
DISEASE & 0.5607 & 0.6608 & 0.6067 & 454\\
RAREDISEASE & 0.8522 & 0.8530 & \bf 0.8526 & 1095\\
SIGN & 0.5574 & 0.5877 & \bf 0.5722 & 958 \\
SYMPTOM & 0.5143 & 0.6667 & 0.5806 & 54 \\

micro avg & 0.6761 & 0.7157 & \bf 0.6954 & 2561\\
macro avg & 0.6212 & 0.6920 & 0.6530 & 2561\\
macro-weighted & 0.6831 & 0.7157 & \bf 0.6984 & 2561\\

\hline

& \multicolumn{1}{c}{\textbf{BioClinical BERT}} \\
{\bf Label} & {\bf Precision} &
{\bf Recall} &
{\bf F1-score} & {\bf Support} \\
\hline\\[-8pt]
DISEASE & 0.5788 & 0.6388 & \bf 0.6073 &   454\\
RAREDISEASE & 0.8167 & 0.8584 & 0.8370 &  1095\\
SIGN & 0.5296 & 0.5501 & 0.5397 &   958\\
SYMPTOM & 0.6066 & 0.6852 & \bf 0.6435 &54\\

micro avg & 0.6625 & 0.7005 & 0.6810 &  2561\\
macro avg & 0.6329 & 0.6831 & \bf 0.6569 &  2561\\

macro-weighted & 0.6627 & 0.7005 & 0.6810 &  2561\\

\hline
\end{tabular}
\end{table}

BioBERT achieves the best micro-average and macro-weighted average F1-scores, while the best macro-average F1-score is provided by ClinicalBERT. In general, BioBERT and ClinicalBERT show very close results. As happened with the previous models, rare diseases show the best results, followed by diseases.
BioBERT obtains the best F1-score for rare diseases and for signs, while ClinicalBERT BERT provides the best results for diseases and symptoms. As expected, the BERT base model obtains lower results than BioBERT and ClinicalBERT.

\section{Conclusions}
\label{sec:conclu}
Although rare diseases have a very low prevalence in the population, more than 400 million people worldwide (around 6\% of the world's population) suffer a rare disease. This number is continually growing as five new rare diseases are discovered each week  \citep{maria2010rare}. 

This work explores different approaches for recognizing rare diseases and their clinical manifestations. We propose a CRF baseline system using linguistic features. Second, we implement several BiLSTMs, exploring different strategies to initialize their input vectors, such as random initialization and three pre-trained word embedding models, one of them was trained on biomedical texts. Moreover, we explore three implementations of BERT, which differ between them by the type of texts used to pre-train the model. The RareDis corpus is used to train the models and evaluate them. The experiments show that BioBERT obtains the best micro and macro-weighted-average F1-score, with  improvements around 5\% over the baseline results. BiLSTM does not even outperform the baseline in terms of F1-score. 
Regarding the entity types, rare diseases show the highest F1-score (85.2\%), while the other entity types do not outperform 60\% in F1-score.

As future work, we plan to extend the size of the RareDis corpus by including MedLine abstracts and clinical cases of rare diseases. This could have a significant positive effect on the results, especially those achieved by the deep learning models. Moreover, we could know if the difference between symptoms and signs is due to their representations or the number of their instances to train the model. We also plan to extend the corpus with texts written in other languages than English. 
We will also address some unsolved problems in NER such as the recognition of nested, overlapped and discontinuous entities.

Regarding the methods, we will study on fine-tuning the BERT-based models by adding CRF to improve the results for signs and symptoms. Furthermore, we plan to address the task of relation extraction on the RareDis corpus.

\section*{Author's contributions}

ISB and SGA conceived and designed the research.
ISB and DCP conducted the literature review.
DCP implemented all methods and performed their experiments. 
All authors discussed the results.
ISB and DCP authors wrote the paper. 
All authors read and approved the final manuscript.

\section*{Funding}
This work was supported by the Madrid Government (Comunidad de Madrid) under the Multiannual Agreement with UC3M in the line of "Fostering Young Doctors Research" (NLP4RARE-CM-UC3M) and in the context of the V PRICIT (Regional Programme of Research and Technological Innovation; the Multiannual Agreement with UC3M in the line of "Excellence of University Professors (EPUC3M17)"; and a grant from Spanish Ministry of Economy and Competitiveness (SAF2017-86810-R).



\newpage
\appendix
\section{Results on token level}
\label{sec:apptoken}
This appendix contains the token-level results of the proposed models  in this study. 
\begin{table}[H]
\centering
\caption{\label{tab:CRFtoken}Token-level results of CRF}
\begin{tabular}{lcccr}
\\
{\bf Label} & {\bf Precision} &
{\bf Recall} &
{\bf F1-score} & {\bf Support} \\
\hline\\[-8pt]

B-DISEASE	&	0.7116	&	0.5124	&	0.5958	&	454\\
I-DISEASE	&	0.7133	&	0.5225	&	0.6032	&	400\\
B-RAREDISEASE	&	0.8464	&	0.8369	&	\bf 0.8416	&	1095\\
I-RAREDISEASE	&	0.8681	&	0.8261	&	\bf 0.8466	&	1179\\
B-SYMPTOM	&	0.8286	&	0.5800	&	0.6824	&	54\\
I-SYMPTOM	&	0.6429	&	0.2250	&	0.3333	&	80\\
B-SIGN	&	0.5883	&	0.4894	&	0.5343	&	958\\
I-SIGN	&	0.5591	&	0.3991	&	0.4658	&	2215\\
\\
micro-avg	&	0.7112	&	0.5818	&	0.6400	&	6243\\
macro-avg	&	0.7198	&	0.5489	&	0.6129	&	6243\\
macro-weighted	&	0.6945	&	0.5818	&	0.6292	&	6243\\
\hline

\end{tabular}
\end{table}

\begin{table}[H]
\centering
\tiny
\caption{\label{tab:tokenBiLSTM}Token-level Results of BiLSTM}
\begin{tabular}{lcccr}
\\
& \multicolumn{1}{c}{\textbf{Random Initialization}} \\
{\bf Label} & {\bf Precision} &
{\bf Recall} &
{\bf F1-score} & {\bf Support} \\
\hline\\[-8pt]
B-DISEASE	&	0.6105	&	0.3102	&	0.4113	&	454\\
I-DISEASE	&	0.6447	&	0.3660	&	0.4669	&	400\\
B-RAREDISEASE	&	0.6232	&	0.5804	&	0.6010	&	1095\\
I-RAREDISEASE	&	0.7812	&	0.6631	&	0.7174	&	1179\\
B-SYMPTOM	&	0.0000	&	0.0000	&	0.0000	&	54\\I-SYMPTOM	&	0.0000	&	0.0000	&	0.0000	&	80\\B-SIGN	&	0.5930	&	0.3311	&	0.4249	&	958\\I-SIGN	&	0.5924	&	0.4323	&	0.4999	&	2215\\

micro-avg	&	0.6403	&	0.4633	&	0.5376	&	6243\\macro-avg	&	0.4806	&	0.3354	&	0.3902	&	6243\\
macro-weighted	&	0.6227	&	0.4633	&	0.5271	&	6243\\

\hline
\\
& \multicolumn{1}{c}{\textbf{GoogleNews}} \\
{\bf Label} & {\bf Precision} &
{\bf Recall} &
{\bf F1-score} & {\bf Support} \\
\hline\\[-8pt]

B-DISEASE	&	0.6301	&	0.3690	&	0.4654	&	454\\
I-DISEASE	&	0.6807	&	0.3256	&	0.4405	&	400\\
B-RAREDISEASE	&	0.6729	&	0.6392	&	0.6556	&	1095\\
I-RAREDISEASE	&	0.8259	&	0.6375	&	0.7196	&	1179\\
B-SYMPTOM	&	0.6452	&	0.4082	&	0.5000	&	54\\
I-SYMPTOM	&	0.5000	&	0.0263	&	0.0500	&	80\\
B-SIGN	&	0.5980	&	0.4178	&	0.4919	&	958\\
I-SIGN	&	0.6203	&	0.4477	&	0.5200	&	2215\\

micro-avg	&	0.6685	&	0.4906	&	0.5659	&	6243\\
macro-avg	&	0.6466	&	0.4089	&	0.4804	&	6243\\
macro-weighted	&	0.6640	&	0.4906	&	0.5593	&	6243\\
\hline

& \multicolumn{1}{c}{\textbf{Glove}} \\
{\bf Label} & {\bf Precision} &
{\bf Recall} &
{\bf F1-score} & {\bf Support} \\
\hline\\[-8pt]
B-DISEASE	&	0.6230	&	0.4198	&	0.5016	&	454\\
I-DISEASE	&	0.6320	&	0.4553	&	0.5293	&	400\\
B-RAREDISEASE	&	0.6838	&	0.6765	&	0.6801	&	1095\\
I-RAREDISEASE	&	0.8321	&	0.6702	&	0.7424	&	1179\\
B-SYMPTOM	&	0.6562	&	0.4286	&	0.5185	&	54\\
I-SYMPTOM	&	0.6667	&	0.1053	&	0.1818	&	80\\
B-SIGN	&	0.5937	&	0.5354	&	0.5630	&	958\\I-SIGN	&	0.5994	&	0.5454	&	0.5711	&	2215\\

micro-avg	&	0.6544	&	0.5683	&	0.6083	&	6243\\
macro-avg	&	0.6609	&	0.4796	&	0.5360	&	6243\\
macro-weighted avg	&	0.6568	&	0.5683	&	0.6059	&	6243\\

\hline
& \multicolumn{1}{c}{\textbf{Wiki-Pubmed-PMC}} \\
{\bf Label} & {\bf Precision} &
{\bf Recall} &
{\bf F1-score} & {\bf Support} \\
\hline\\[-8pt]
B-DISEASE	&	0.7600	&	0.4718	&	0.5822	&	454\\
I-DISEASE	&	0.7546	&	0.5150	&	0.6122	&	400\\
B-RAREDISEASE	&	0.7163	&	0.6636	&	0.6889	&	1095\\
I-RAREDISEASE	&	0.8489	&	0.6480	&	0.7350	&	1179\\B-SYMPTOM	&	0.6765	&	0.4600	&	0.5476	&	54\\I-SYMPTOM	&	1.0000	&	0.0750	&	0.1395	&	80\\
B-SIGN	&	0.5318	&	0.5106	&	0.5210	&	958\\
I-SIGN	&	0.5807	&	0.4614	&	0.5142	&	2215\\micro-avg	&	0.6687	&	0.5369	&	0.5956	&	6243\\macro-avg	&	0.7336	&	0.4757	&	0.5426	&	6243\\
macro-weighted avg	&	0.6784	&	0.5369	&	0.5934	&	6243\\
\hline
\end{tabular}
\end{table}

\begin{table}[H]
\centering
\tiny
\caption{\label{tab:tokenBiLSTMCRF}Token-level results of BiLSTM+CRF models}
\begin{tabular}{lcccr}
\\
& \multicolumn{1}{c}{\textbf{Random Initialization}} \\
{\bf Label} & {\bf Precision} &
{\bf Recall} &
{\bf F1-score} & {\bf Support} \\
\hline\\[-8pt]

B-DISEASE	&	0.5714	&	0.3957	&	0.4676	&	454\\
I-DISEASE	&	0.5649	&	0.4640	&	0.5095	&	400\\
B-RAREDISEASE	&	0.6858	&	0.7490	&	0.7160	&	1095\\
I-RAREDISEASE	&	0.7703	&	0.7710	&	0.7707	&	1179\\
B-SYMPTOM	&	0.9375	&	0.6122	&	0.7407	&	54\\
I-SYMPTOM	&	0.8333	&	0.2632	&	0.4000	&	80\\
B-SIGN	&	0.6029	&	0.5616	&	0.5816	&	958\\
I-SIGN	&	0.6112	&	0.5669	&	0.5882	&	2215\\
micro-avg	&	0.6521	&	0.6118	&	0.6313	&	6243\\
macro-avg	&	0.6972	&	0.5480	&	0.5968	&	6243\\
macro-weighted	&	0.6499	&	0.6118	&	0.6270	&	6243\\

\hline
\\
& \multicolumn{1}{c}{\textbf{GoogleNews}} \\
{\bf Label} & {\bf Precision} &
{\bf Recall} &
{\bf F1-score} & {\bf Support} \\
\hline\\[-8pt]

B-DISEASE	&	0.6123	&	0.4519	&	0.5200	&	454\\
I-DISEASE	&	0.5953	&	0.5130	&	0.5511	&	400\\
B-RAREDISEASE	&	0.6913	&	0.7990	&	0.7412	&	1095\\
I-RAREDISEASE	&	0.7727	&	0.8117	&	0.7917	&	1179\\
B-SYMPTOM	&	0.8108	&	0.6122	&	0.6977	&	54\\
I-SYMPTOM	&	0.6818	&	0.1974	&	0.3061	&	80\\
B-SIGN	&	0.6624	&	0.5308	&	0.5894	&	958\\
I-SIGN	&	0.7074	&	0.5236	&	0.6018	&	2215\\
micro-avg	&	0.7022	&	0.6103	&	0.6530	&	6243\\
macro-avg	&	0.6918	&	0.5549	&	0.5999	&	6243\\
macro-weighted	&	0.6992	&	0.6103	&	0.6450	&	6243\\

\hline

& \multicolumn{1}{c}{\textbf{Glove}} \\
{\bf Label} & {\bf Precision} &
{\bf Recall} &
{\bf F1-score} & {\bf Support} \\
\hline\\[-8pt]
B-DISEASE	&	0.5219	&	0.5428	&	0.5321	&	454\\
I-DISEASE	&	0.4875	&	0.6167	&	0.5445	&	400\\
B-RAREDISEASE	&	0.7792	&	0.7510	&	0.7649	&	1095\\
I-RAREDISEASE	&	0.8009	&	0.8037	&	0.8023	&	1179\\
B-SYMPTOM	&	0.6739	&	0.6327	&	0.6526	&	54\\
I-SYMPTOM	&	0.4878	&	0.2632	&	0.3419	&	80\\
B-SIGN	&	0.6372	&	0.5753	&	0.6047	&	958\\
I-SIGN	&	0.6566	&	0.5730	&	0.6120	&	2215
\\micro-avg	&	0.6789	&	0.6390	&	0.6583	&	6243\\macro-avg	&	0.6306	&	0.5948	&	0.6069	&	6243\\
macro-weighted	&	0.6798	&	0.6390	&	0.6572	&	6243\\

\hline
& \multicolumn{1}{c}{\textbf{Wiki-Pubmed-PMC}} \\
{\bf Label} & {\bf Precision} &
{\bf Recall} &
{\bf F1-score} & {\bf Support} \\
\hline\\[-8pt]
B-DISEASE	&	0.7616	&	0.5192	&	0.6174	&	454\\
I-DISEASE	&	0.7789	&	0.5550	&	0.6482	&	400\\
B-RAREDISEASE	&	0.6617	&	0.8295	&	0.7361	&	1095\\
I-RAREDISEASE	&	0.7694	&	0.8346	&	0.8007	&	1179\\
B-SYMPTOM	&	0.7273	&	0.6400	&	0.6809	&	54\\
I-SYMPTOM	&	0.6296	&	0.2125	&	0.3178	&	80\\
B-SIGN	&	0.5919	&	0.6015	&	0.5967	&	958\\
I-SIGN	&	0.5929	&	0.5589	&	0.5754	&	2215\\
micro-avg	&	0.6621	&	0.6561	&	0.6591	&	6243\\
macro-avg	&	0.6892	&	0.5939	&	0.6216	&	6243\\
macro-weighted	&	0.6634	&	0.6561	&	0.6535	&	6243\\
\hline
\end{tabular}
\end{table}

\begin{table}[H]
\centering
\tiny
\caption{\label{tab:Tokenbert}Token-level results of the BERT-based models}
\begin{tabular}{lcccc}
\\
& \multicolumn{1}{c}{\textbf{BERT base}} \\
{\bf Label} & {\bf Precision} &
{\bf Recall} &
{\bf F1-score} & {\bf Support} \\
\hline\\[-8pt]
 B-DISEASE & 0.6012 & 0.6637 & 0.6309 &   454\\
  I-DISEASE & 0.5186 & 0.5884 & 0.5513 &  400\\
B-RAREDISEASE & 0.8451 & 0.9003 & 0.8718 &  1095\\
I-RAREDISEASE & 0.8704 & 0.9024 & 0.8861 &  1179\\
  B-SYMPTOM & 0.6607 & 0.7400 & 0.6981 &    54\\
  I-SYMPTOM & 0.6000 & 0.4918 & 0.5405 &   80\\
    B-SIGN & 0.6514 & 0.7073 & 0.6782 &   958\\
       I-SIGN & 0.6725 & 0.7099 & 0.6907 &  2215\\

    micro avg & 0.7353 & 0.7794 & 0.7567 & 6243\\
    macro avg & 0.6775 & 0.7130 & 0.6935 & 6243\\
 macro-weighted avg & 0.7379 & 0.7794& 0.7579 & 6243\\

\hline

& \multicolumn{1}{c}{\textbf{BioBERT}} \\
{\bf Label} & {\bf Precision} &
{\bf Recall} &
{\bf F1-score} & {\bf Support} \\
\hline\\[-8pt]
B-DISEASE & 0.6356 & 0.7088 & 0.6702 &   454\\
I-DISEASE & 0.5716 & 0.6964 & 0.6279 &  400\\
    
B-RAREDISEASE & 0.8825 & 0.8816 & 0.8821 &  1095\\
I-RAREDISEASE & 0.9142 & 0.8927 & 0.9033 &  1179\\
    
B-SYMPTOM & 0.6349 & 0.8000 & 0.7080 &    54\\
I-SYMPTOM & 0.5538 & 0.5538 & 0.5538 &   80\\

   B-SIGN & 0.7238 & 0.7049 & 0.7142 &   958\\
    I-SIGN & 0.7330 & 0.6978 & 0.7150 &  2215\\
    
    micro avg & 0.7830 & 0.7855 & 0.7842 & 6243\\
    macro avg & 0.7062 & 0.7420 & 0.7218 & 6243\\
 macro-weighted avg & 0.7890 & 0.7855 & 0.7863 & 6243\\

\hline

& \multicolumn{1}{c}{\textbf{ClinicalBERT}} \\
{\bf Label} & {\bf Precision} &
{\bf Recall} &
{\bf F1-score} & {\bf Support} \\
\hline\\[-8pt]
B-DISEASE & 0.6503 & 0.6885 & 0.6689 &   454\\
I-DISEASE & 0.5969 & 0.6557 & 0.6249 &  400\\
B-RAREDISEASE & 0.8614 & 0.8807 & 0.8710 &  1095\\
I-RAREDISEASE & 0.8829 & 0.9076 & 0.8951 &  1179\\
    B-SYMPTOM & 0.7547 & 0.8000 & 0.7767 &    54\\
    I-SYMPTOM & 0.7158 & 0.5231 & 0.6044 &   80\\
       B-SIGN & 0.6996 & 0.6961 & 0.6979 &   958\\
       I-SIGN & 0.7575 & 0.6220 & 0.6831 &  2215\\

    micro avg & 0.7881 & 0.7609 & 0.7742 & 6243\\
    macro avg & 0.7399 & 0.7217 & 0.7277 & 6243\\
 macro-weighted avg & 0.7873 & 0.7609 & 0.6243 & 11909\\

\hline
\end{tabular}
\end{table}



\end{document}